\documentclass[runningheads]{llncs}
\usepackage{graphicx}
\usepackage{comment}
\usepackage{amsmath,amssymb} 
\usepackage{color}

\usepackage{siunitx}
\usepackage{cite}
\usepackage{ctable}
\usepackage{hhline}
\usepackage{booktabs}
\usepackage{csquotes}
\usepackage{pifont}
\usepackage{multirow}
\usepackage{booktabs}
\usepackage{capt-of}
\usepackage{subcaption}

\newcolumntype{L}[1]{>{\raggedright\let\newline\\\arraybackslash\hspace{0pt}}m{#1}}
\newcolumntype{C}[1]{>{\centering\let\newline\\\arraybackslash\hspace{0pt}}m{#1}}
\newcolumntype{R}[1]{>{\raggedleft\let\newline\\\arraybackslash\hspace{0pt}}m{#1}}
\newcommand{\cmark}{\ding{51}}%
\usepackage[pagebackref=true,breaklinks=true,letterpaper=true,colorlinks,bookmarks=false]{hyperref}

\definecolor{brown}{rgb}{0.65, 0.16, 0.16}
\definecolor{gray}{rgb}{0.5,0.5,0.5}

\newcommand*{\affmark}[1][*]{\textsuperscript{#1}}
\usepackage{wrapfig}
\usepackage[symbol]{footmisc}

\def\eg{\emph{e.g.}}
\def\ie{\emph{i.e.}}

\begin{document}

\pagestyle{headings}
\mainmatter
\def\ECCVSubNumber{2553}  

\title{MotionSqueeze: Neural Motion Feature Learning for Video Understanding}



\author{
Heeseung Kwon\affmark[1,2] \hspace{0.4cm}
Manjin Kim\affmark[1]  \hspace{0.4cm}
Suha Kwak\affmark[1]  \hspace{0.4cm}
Minsu Cho\affmark[1,2]
}

\institute{\affmark[1]POSTECH\footnotemark[1]\hspace{2.0cm}
\affmark[2]NPRC\footnotemark[2]\\
{\tt\small \url{http://cvlab.postech.ac.kr/research/MotionSqueeze/}}
}

\authorrunning{Heeseung Kwon, Manjin Kim, Suha Kwak, and Minsu Cho}

\maketitle


\begin{abstract}

Motion plays a crucial role in understanding videos and most state-of-the-art neural models for video classification incorporate motion information typically using optical flows extracted by a separate off-the-shelf method. As the frame-by-frame optical flows require heavy computation, incorporating motion information has remained a major computational bottleneck for video understanding. In this work, we replace external and heavy computation of optical flows with internal and light-weight learning of motion features.  
We propose a trainable neural module, dubbed {\em MotionSqueeze}, for effective motion feature extraction. 
Inserted in the middle of any neural network, it learns to establish correspondences across frames and convert them into motion features, which are readily fed to the next downstream layer for better prediction. We demonstrate that the proposed method provides a significant gain on four standard benchmarks for action recognition with only a small amount of additional cost, outperforming the state of the art on Something-Something-V1\&V2 datasets.

\end{abstract}

\footnotetext[1]{Pohang University of Science and Technology, Pohang, Korea}
\footnotetext[2]{The Neural Processing Research Center, Seoul, Korea}


\section{Introduction}
The most distinctive feature of videos, from those of images, is motion. 
In order to grasp a full understanding of a video, we need to analyze its motion patterns as well as the appearance of objects and scenes in the video~\cite{wang2011action,simonyan2014two,lee2018motion,piergiovanni2019representation}.
With significant progress of neural networks on the image domain, convolutional neural networks (CNNs) have been widely used to learn appearance features from video frames~\cite{simonyan2014two,wang2016temporal,tran2015learning,donahue2015long} and recently extended to learn temporal features using spatio-temporal convolution across multiple frames~\cite{carreira2017quo,tran2015learning}. 
The results, however, have shown that spatio-temporal convolution alone is not enough for learning motion patterns;
convolution is effective in capturing translation-equivariant patterns but not in modeling relative movement of objects~\cite{zhao2018trajectory,wang2015action}.
As a result, most state-of-the-art methods still incorporate explicit motion features, {\em i.e.,} dense optical flows, extracted by an external off-the-shelf methods~\cite{simonyan2014two,carreira2017quo,xie2018rethinking,tran2018closer,lin2019tsm}.
This causes a major computational bottleneck in video-processing models for two reasons.
First, calculating optical flows frame-by-frame is a time-consuming process; obtaining optical flows of a video is typically an order of magnitude slower than feed-forwarding the video through a deep neural network.
Second, processing optical flows often requires a separate stream in the model to learn motion representations~\cite{simonyan2014two}, which results in doubling the number of parameters and the computational cost. 
To address these issues, several methods have attempted to internalize motion modeling~\cite{sun2018optical,lee2018motion,fan2018end,piergiovanni2019representation}.
They, however, all either impose a heavy computation on their architectures~\cite{fan2018end,piergiovanni2019representation} or underperform other methods using external optical flows~\cite{sun2018optical,lee2018motion}.

\begin{figure}[t]
\centering
\includegraphics[width=0.77\columnwidth]{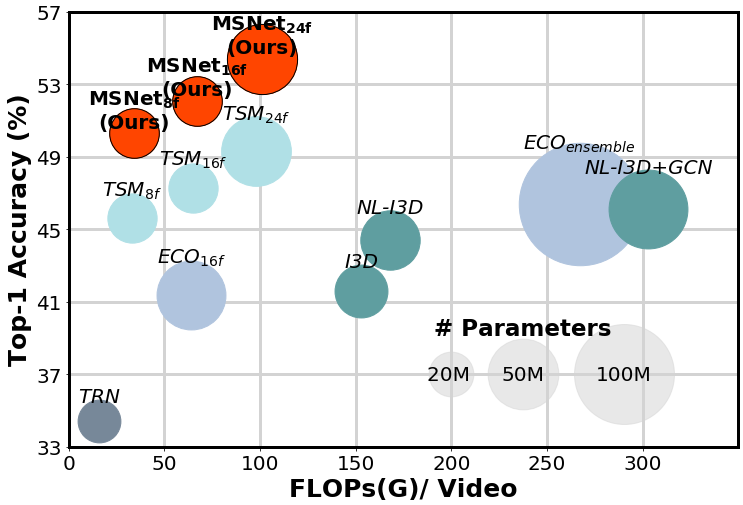}
   \caption{
   Video classification performance comparison on Something-Something-V1~\cite{goyal2017something} in terms of accuracy, computational cost, and model size. The proposed method (MSNet) achieves the best trade-off between accuracy and efficiency compared to state-of-the-art methods of TSM~\cite{lin2019tsm}, TRN~\cite{zhou2018temporal}, ECO~\cite{zolfaghari2018eco}, I3D~\cite{carreira2017quo}, NL-I3D~\cite{wang2018non}, and GCN~\cite{wang2018videos}. Best viewed in color. 
   }
\label{fig:1}
\end{figure}

To tackle the limitation of the existing methods, we propose an end-to-end trainable block, dubbed the {\em MotionSqueeze} (MS) module, for effective motion estimation.
Inserted in the middle of any neural network for video understanding, it learns to establish correspondences across adjacent frames efficiently and convert them into effective motion features. The resultant motion features are readily fed to the next downstream layer and used for final prediction.   
To validate the proposed MS module, we develop a video classification architecture, dubbed the MotionSqueeze network (MSNet), that is equipped with the MS module. 
In comparison with recent methods, shown in Figure~\ref{fig:1}, the proposed method provides the best trade-off in terms of accuracy, computational cost, and model size in video understanding.


\section{Related work}

\noindent
\textbf{Video classification architectures.}
One of the main problems in video understanding is to categorize videos given a set of pre-defined target classes.
Early methods based on deep neural networks have focused on learning spatio-temporal or motion features. Tran~\emph{et al.}~\cite{tran2015learning} propose a 3D CNN (C3D) to learn spatio-temporal features while Simonyan and Zisserman~\cite{simonyan2014two} employ an independent temporal stream to learn motion features from precomputed optical flows.
Carreira and Zisserman~\cite{carreira2017quo} design two-stream 3D CNNs (two-stream I3D) by integrating two former methods, and achieve the state-of-the-art performance at that time.
As the two-stream 3D CNNs are powerful but computationally demanding, subsequent work has attempted to improve the efficiency.
Tran~\emph{et al.}\cite{tran2018closer} and  Xie~\emph{et al.}\cite{xie2018rethinking} propose to decompose 3D convolutional filters into 2D spatial and 1D temporal filters.
Chen~\emph{et al.}~\cite{chen2018multi} adopt group convolution techniques while Zolfaghari~\emph{et al.}~\cite{zolfaghari2018eco} propose to study mixed 2D and 3D networks with the frame sampling method of temporal segment networks (TSN)~\cite{wang2016temporal}.
Tran~\emph{et al.}~\cite{tran2019video} analyze the effect of 3D group convolutional networks and propose the channel-separated convolutional network (CSN).
Lin~\emph{et al.}~\cite{lin2019tsm} propose the temporal shift module (TSM) that simulates 3D convolution using 2D convolution with a part of input feature channels shifted along the temporal axis. It enables 2D convolution networks to achieve a comparable classification accuracy to 3D CNNs. Unlike these approaches, we focus on efficient learning of motion features. 

\smallbreak
\noindent
\textbf{Learning motions in a video.}
While two-stream-based architectures~\cite{simonyan2014two,feichtenhofer2016convolutional,feichtenhofer2016spatiotemporal,sevilla2018integration} have demonstrated the effectiveness of pre-computed optical flows, the use of optical flows typically degrades the efficiency of video processing.
To address the issue, Ng~\emph{et al.}~\cite{ng2018actionflownet} use a multi-task learning of both optical flow estimation and action classification and Stroud~\emph{et al.}~\cite{stroud2020d3d} propose to distill motion features from pre-trained two-stream 3D CNNs.
These methods do not use pre-computed optical flows during inference, but still need them at the training phase.
Other methods design network architectures that learn motions internally without external optical flows~\cite{sun2018optical,jiang2019stm,lee2018motion,fan2018end,piergiovanni2019representation}.
Sun~\emph{et al.}~\cite{sun2018optical} compute spatial and temporal gradients between appearance features to learn motion features.
Lee~\emph{et al.}~\cite{lee2018motion} and Jiang~\emph{et al.}~\cite{jiang2019stm} propose a convolutional module to extract motion features by spatial shift and subtraction operation between appearance features.
Despite their computational efficiency, they do not reach the classification accuracy of two-stream networks~\cite{simonyan2014two}.
Fan~\emph{et al.}~\cite{fan2018end} implement the optimization process of TV-L1~\cite{zach2007duality} as iterative neural layers, and design an end-to-end trainable architecture (TVNet).
Piergiovanni and Ryoo~\cite{piergiovanni2019representation} extend the idea of TVNet by calculating channel-wise flows of feature maps at the intermediate layers of the CNN.
These variational methods achieve a good performance, but require a high computational cost due to iterative neural layers.
In contrast, our method learns to extract effective motion features with a marginal increase of computation.

\smallbreak
\noindent
\textbf{Learning visual correspondences.}
Our work is inspired by recent methods that learn visual correspondences between images using neural networks~\cite{han2017scnet,dosovitskiy2015flownet,rocco2017convolutional,min2019hyperpixel,lee2019sfnet,sun2018pwc}.
Fischer~\emph{et al.}~\cite{dosovitskiy2015flownet} estimate optical flows using a convolutional neural network, which construct a correlation tensor from feature maps and regresses displacements from it. Sun~\emph{et al.}~\cite{sun2018pwc} use a stack of correlation layers for coarse-to-fine optical flow estimation.
While these methods require dense ground-truth optical flows in training, 
the structure of correlation computation and subsequent displacement estimation is widely adopted in other correspondence problems with different levels of supervision. For example, recent methods for semantic correspondence, \ie, matching images with intra-class variation, typically follow a similar pipeline to learn geometric transformation between images in a more weakly-supervised regime~\cite{han2017scnet,rocco2017convolutional,lee2019sfnet,min2019hyperpixel}. In this work, motivated by this line of research, we develop a motion feature module that does not require any correspondence supervision for learning.  

Similarly to our work, a few recent methods~\cite{liu2019learning,zhao2018recognize} have attempted to incorporate correspondence information for video understanding.
Zhao~\emph{et al.}~\cite{zhao2018recognize} use correlation information between feature maps of consecutive frames to replace optical flows. The size of their full model, however, is comparable to the two-stream networks~\cite{simonyan2014two}.
Liu~\emph{et al.}~\cite{liu2019learning} propose the correspondences proposal (CP) module to learn correspondences in a video. 
Unlike ours, they focus on analyzing spatio-temporal relationship within the whole video, rather than motion, and the model is not fully differentiable and thus less effective in learning.
In contrast, we introduce a fully-differentiable motion feature module that can be inserted in the middle of any neural network for video understanding. 

The main contribution of this work is three-fold.
\begin{itemize}
\item We propose an end-to-end trainable, model-agnostic, and lightweight module for motion feature extraction. 
\item We develop an efficient video recognition architecture that is equipped with the proposed motion module.
\item We demonstrate the effectiveness of our method on four different benchmark datasets and achieve the state-of-the-art on Something-Something-V1\&V2.
\end{itemize}


\begin{figure*}[t]
    \centering
    \includegraphics[width=\columnwidth]{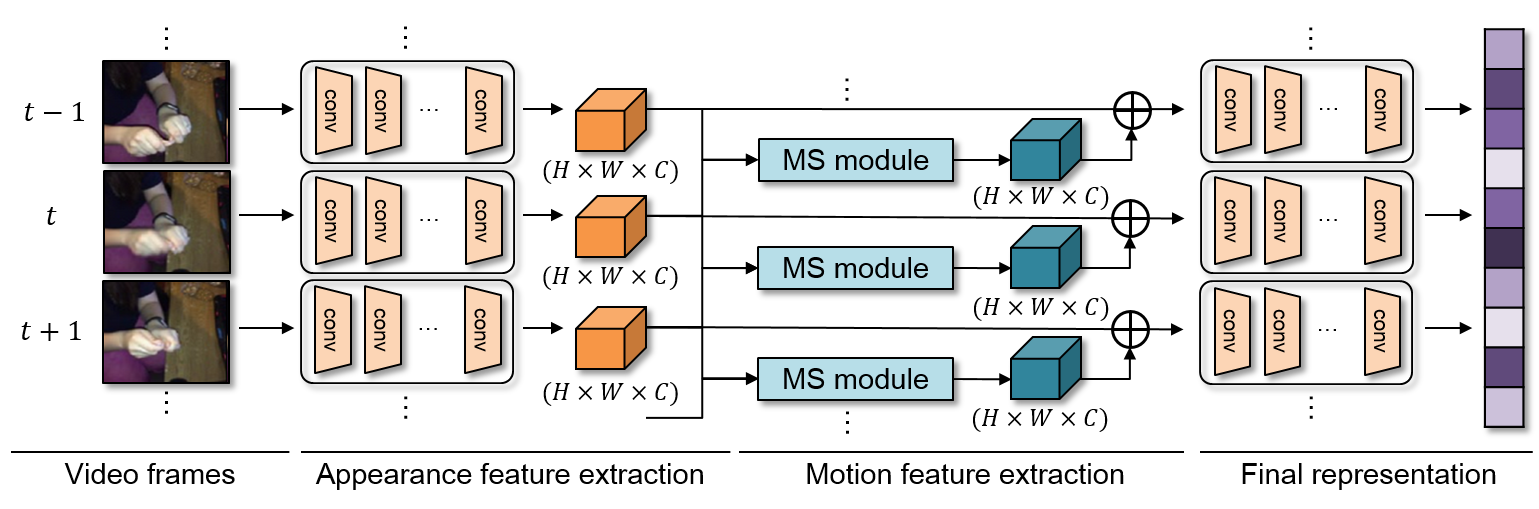}
\caption{Overall architecture of the proposed approach. The model first takes $T$ video frames as input and converts them into frame-wise appearance features using convolutional layers. The proposed \emph{MotionSqueeze (MS) module} generates motion features using the frame-wise appearance features, and combines the motion features into the next downstream layer. $\oplus$ denotes element-wise addition. 
} \label{fig:2}
\end{figure*}

\section{Proposed approach}
The overall architecture for video understanding is illustrated in Figure~\ref{fig:2}.
Let us assume a neural network that takes a video of $T$ frames as input and predicts the category of the video as output, where convolutional layers are used to transform input frames into frame-wise appearance features. The proposed motion feature module, dubbed {\em MotionSqueeze (MS) module}, is inserted to produce frame-wise motion features using pairs of adjacent appearance features. The resultant motion features are added to the appearance features for final prediction.
In this section, we first explain the MS module, and describe the details of our network architecture for video understanding.

\subsection{MotionSqueeze (MS) module}
\begin{figure*}[t]
    \centering
    \includegraphics[width=\columnwidth]{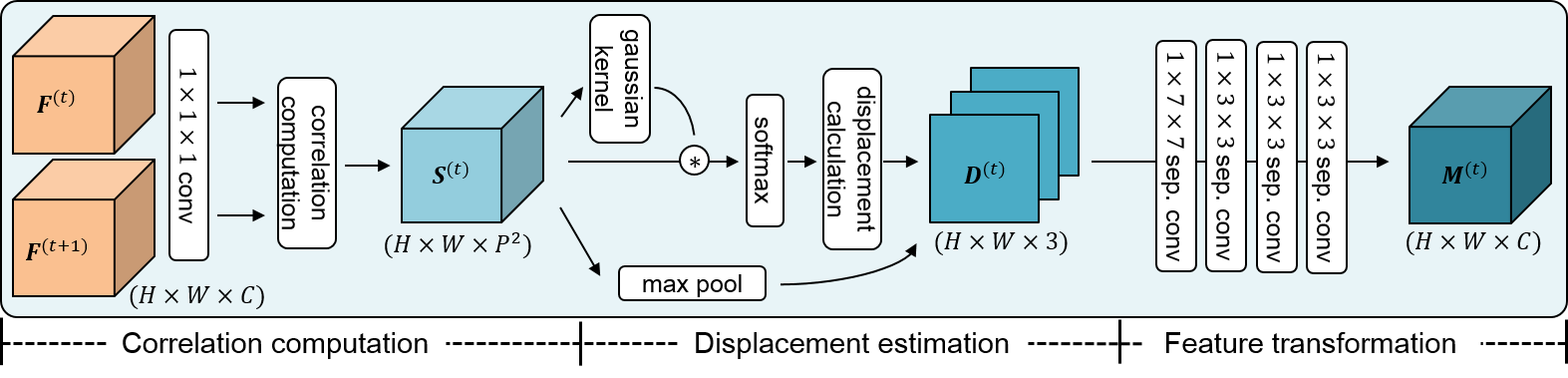}
\caption{
Overall process of MotionSqueeze (MS) module. The MS module estimates motion across two frame-wise feature maps ($\mathbf{F}^{(t)}, \mathbf{F}^{(t+1)}$) of adjacent frames. A correlation tensor $\mathbf{S}^{(t)}$ is obtained by computing correlations, and then a displacement tensor $\mathbf{D}^{(t)}$ is estimated using the tensor. Through the transformation process of convolution layers, the final motion feature $\mathbf{M}^{(t)}$ is obtained.
See text for details.
} \label{fig:3}
\end{figure*}

The MS module is a learnable motion feature extractor, which can replace the use of explicit optical flows for video understanding.
As described in Figure~\ref{fig:3}, given two feature maps from adjacent frames, it learns to extract effective motion features in three steps: correlation computation, displacement estimation, and feature transformation. 

\smallbreak
\noindent
\textbf{Correlation computation.} 
Let us denote the two adjacent input feature maps by $\mathbf{F}^{(t)}$ and $\mathbf{F}^{(t+1)}$, each of which is 3D tensors of size $H \times W \times C$. The spatial resolution is $H \times W$ and the $C$ dimensional features on spatial position $\mathbf{x}$ by $\mathbf{F}_{\mathbf{x}}$.
A correlation score of position $\mathbf{x}$ with respect to displacement $\mathbf{p}$ is defined as 
\begin{align}
    s(\mathbf{x},\mathbf{p},t)=\mathbf{F}^{(t)}_{\mathbf{x}} \cdot \mathbf{F}^{(t+1)}_{\mathbf{x}+\mathbf{p}},
\end{align}
where $\cdot$ denotes dot product.
For efficiency, we compute the correlation scores of position $\mathbf{x}$ only in its neighborhood of size $P=2k+1$ by restricting a maximum displacement:  $\mathbf{p}\in[-k,k]^2$.
For $t_{\rm th}$ frame, a resultant correlation tensor $\mathbf{S}^{(t)}$ is of size $H \times W \times P^2$. 
The cost of computing the correlation tensor is equivalent to that of  $1\times1$ convolutions with $P^2$ kernels; the correlation computation can be implemented as 2D convolutions on $t_{\rm th}$ feature map using $t+1_{\rm th}$ feature map as $P^2$ kernels.
The total FLOPs in a single video amounts to $T H W C P^2$.
We apply a convolution layer before computing correlations, which learns to weight informative feature channels for learning visual correspondences.
In practice, we set the neighborhood $P=15$ given the spatial resolution $28\times 28$ and apply an $1\times1\times1$ layer with $C/2$ channels. For correlation computation, we adopt C++/Cuda implemented version of correlation layer in FlowNet~\cite{dosovitskiy2015flownet}.

\smallbreak
\noindent
\textbf{Displacement estimation.} 
From the correlation tensor $\mathbf{S}^{(t)}$, we estimate a displacement field for motion information.
A straightforward but non-differentiable method would be to take the best matching displacement for position $\mathbf{x}$ by $\mathrm{argmax}_{\mathbf{p}}s(\mathbf{x},\mathbf{p},t)$. 
To make the operation differentiable, we can use a weighted average of displacements using softmax, called {\em soft-argmax}~\cite{honari2018improving,lee2019sfnet}, which is defined as 
\begin{align}
    d(\mathbf{x},t) = \sum_{\mathbf{p}} 
    \frac{\exp(s(\mathbf{x},\mathbf{p},t))}{\sum_{\mathbf{p}'}{\exp(s(\mathbf{x},\mathbf{p}',t))}} \mathbf{p}. 
\end{align}
This method, however, is sensitive to noisy outliers in the correlation tensor since it is influenced by all correlation values.
We thus use the {\em kernel-soft-argmax}~\cite{lee2019sfnet} that suppresses such outliers by masking a 2D Gaussian kernel on the correlation values; 
the kernel is centered on each target position so that the estimation is more influenced by closer neighbors.
Our kernel-soft-argmax for displacement estimation is defined as 
\begin{align}
    d(\mathbf{x},t) = \sum_{\mathbf{p}}
    \frac{\exp(g(\mathbf{x},\mathbf{p},t)s(\mathbf{x},\mathbf{p},t) / \tau )}{\sum_{\mathbf{p}'}{\exp( g(\mathbf{x},\mathbf{p}',t) s(\mathbf{x},\mathbf{p}',t) / \tau )}}  \mathbf{p},
\end{align}
where
\begin{align}
    g(\mathbf{x},\mathbf{p},t) = \frac{1}{\sqrt{2\pi}\sigma}\exp(\frac{\mathbf{p}-\mathrm{argmax}_{\mathbf{p}}s(\mathbf{x},\mathbf{p},t)}{\sigma^{2}}).
\end{align}
Note that $g(\mathbf{x},\mathbf{p},t)$ is the Gaussian kernel and we empirically set the standard deviation $\sigma$ to 5.
$\tau$ is a temperature factor adjusting the softmax distribution; as $\tau$ decreases, softmax approaches argmax.
We set $\tau=0.01$ in our experiments.

In addition to the estimated displacement map, we use a confidence map of correlation as auxiliary motion information, which is obtained by pooling the highest correlation on each position $\mathbf{x}$: 
\begin{align}
s^{*}(\mathbf{x},t) =\max_{\mathbf{p}}s(\mathbf{x},\mathbf{p},t). 
\end{align}
The confidence map may be useful for identifying displacement outliers and learning informative motion features.

We concatenate the (2-channel) displacement map and the (1-channel) confidence map into a displacement tensor $\mathbf{D}^{(t)}$ of size $H\times W\times 3$ for the next step of motion feature transformation. An example of them is visualized in Figure~\ref{fig:4}.

\smallbreak
\noindent
\textbf{Feature transformation.}
We convert the displacement tensor $\mathbf{D}^{(t)}$ to an effective motion feature $\mathbf{M}^{(t)}$ that is readily incorporated into downstream layers.
The tensor $\mathbf{D}^{(t)}$ is fed to four \textit{depth-wise separable convolution}~\cite{howard2017mobilenets} layers, one $1\times7\times7$ layer followed by three $1\times3\times3$ layers, and transformed into a motion feature $\mathbf{M}^{(t)}$ with the same number of channels $C$ as that of the original input $\mathbf{F}^{(t)}$.
The depth-wise separable convolution approximates 2D convolution with a significantly less computational cost~\cite{chollet2017xception,sandler2018mobilenetv2,tran2019video}.
Note that all depth-wise and point-wise convolution layers are followed by batch normalization~\cite{ioffe2015batch} and ReLU~\cite{nair2010rectified}.
As in the temporal stream layers of~\cite{simonyan2014two}, this feature transformation process is designed to learn task-specific motion features with convolution layers by interpreting the semantics of displacement and confidence.
As illustrated in Figure~\ref{fig:2}, the MS module generates motion feature $\mathbf{M}^{(t)}$ using two adjacent appearance features $\mathbf{F}^{(t)}$ and $\mathbf{F}^{(t+1)}$ and then add it to the input of the next layer. Given $T$ frames, we simply pad the final motion feature $\mathbf{M}^{(T)}$ with $\mathbf{M}^{(T-1)}$ by setting $\mathbf{M}^{(T)}=\mathbf{M}^{(T-1)}$. 

\subsection{MotionSqueeze network (MSNet).} 
The MS module can be inserted into any video understanding architecture to improve the performance by motion feature modeling. 
In this work, we introduce standard convolutional neural networks (CNNs) with the MS module, dubbed {\em MSNet}, for video classification.  
We adopt the ImageNet-pretrained ResNet~\cite{he2016deep} as the CNN backbone and insert TSM~\cite{lin2019tsm} for each residual block of the ResNet.
TSM enables 2D convolution to obtain the effect of 3D convolution by shifting a part of input feature channels along the temporal axis before the convolution operation.
Following the default setting in~\cite{lin2019tsm}, we shift $1/8$ of the input features channels forward and another $1/8$ of the channels backward in each TSM.

The overall architecture of the proposed model is shown in Figure~\ref{fig:2}; 
a single MS module is inserted after the third stage of the ResNet.
We fuse the motion feature into the appearance feature by element-wise addition:
\begin{align}
\mathbf{F}'^{(t)} =\mathbf{F}^{(t)} + \mathbf{M}^{(t)}.
\end{align}
In section~\ref{ablation}, we extensively evaluate different fusion methods, \eg, concatenation and multiplication, and show that additive fusion is better than the others.
After fusing both features, the combined feature is passed through the next downstream layers.
The network outputs over $T$ frames are temporally averaged to produce a final output and the cross-entropy with softmax is used as a loss function for training.
By default setting, MSNet learns both appearance and motion features jointly in a single network at the cost of only 2.5\% and 1.2\% increase in FLOPs and the number of parameters, respectively.

\section{Experiments}
\subsection{Datasets}
\textbf{Something-Something V1\&V2}~\cite{goyal2017something} are trimmed video datasets for human action classification.
Both datasets consist of 174 classes with 108,499 and 220,847 videos in total, respectively.
Each video contains one action and the duration spans from 2 to 6 seconds.
Something-Something V1\&V2 are motion-oriented datasets where temporal relationships are more salient than in others.

\smallbreak
\noindent
\textbf{Kinetics}~\cite{kay2017kinetics} is a popular large-scale video dataset, consisting of 400 classes with over 250,000 videos. 
Each video lasts around 10 seconds with a single action.

\noindent
\textbf{HMDB51}~\cite{kuehne2011hmdb} contains 51 classes with 6,766 videos.
Kinetics and HMDB-51 focus more on appearance information rather than motion.

\subsection{Implementation details}

\textbf{Clip sampling.} In both training and testing, instead of an entire video, we use a clip of frames that are sampled from the video. 
We use the segment-based sampling method~\cite{wang2016temporal} for the Something-Something V1\&V2 while adopting the dense frame sampling method~\cite{carreira2017quo} for Kinetics and HMDB-51. 

\smallbreak
\noindent
\textbf{Training.}
For each video, we sample a clip of $8$ or $16$ frames, resize them into  $240\times320$ images, and crop $224\times224$ images from the resized images~\cite{zolfaghari2018eco}.
The minibatch SGD with Nestrov momentum is used for optimization, and the batch size is set to 48.
We use scale jittering for data augmentation.
For the Something-Something V1\&V2, we set the training epochs to 40 and the initial learning rate to 0.01; the learning rate is decayed by 1/10 after $20_{\rm th}$ and $30_{\rm th}$ epochs.
For Kinetics, we set the training epochs to 80 and the initial learning rate to 0.01; the learning rate is decayed by 1/10 after 40 and 60 epochs. 
In training our model on HMDB-51, we fine-tune the Kinetics-pretrained model as in~\cite{tran2018closer,lin2019tsm}.
We set the training epochs to 35 and the initial learning rate to 0.001; the learning rate is decayed by 1/10 after $15_{\rm th}$ and $30_{\rm th}$ epochs.

\smallbreak
\noindent
\textbf{Inference.}
Given a video, we sample a clip and test its center crop.
For Something-Something V1\&V2, we evaluate both the single clip prediction and the average prediction of 10 randomly-sampled clips.
For Kinetics and HMDB-51, we evaluate the average prediction of uniformly-sampled 10 clips from each video.

\begin{table}[h]
\centering
\captionsetup{width=0.91\textwidth}
\caption{Performance comparison on Something-Something V1\&V2.
The symbol ${\dagger}$ denotes the reproduced by ours.
\label{sota_table}}
\renewcommand{\arraystretch}{0.9}
\fontsize{8pt}{9.5pt}\selectfont
\scalebox{1.0}{
\begin{tabular}{lcccccccc}
\toprule
model & flow & \#frame  & FLOPs	& \#param & \multicolumn{2}{c}{SomethingV1} & \multicolumn{2}{c}{SomethingV2}	\\
   &    &  & $\times$clips  & & top-1 & top-5  & top-1 & top-5 \\
\midrule
TSN~\cite{wang2016temporal} &		& 8 & 16G$\times$1 & 10.7M & 19.5 & - & 33.4 & - \\
TRN~\cite{zhou2018temporal} &		& 8  & 16G$\times$N/A & 18.3M & 34.4 & - & 48.8 & - \\
TRN Two-stream~\cite{zhou2018temporal} &	\cmark	& 8+8  & 16G$\times$N/A & 18.3M & 42.0 & - & 55.5 & - \\
MFNet~\cite{lee2018motion} &		& 10  & N/A$\times$10 & - & 43.9 & 73.1 & - & - \\
CPNet~\cite{liu2019learning} &		& 24  & N/A$\times$96 & - & - & - & 57.7 & 84.0 \\
\midrule
ECO$_{En}Lite$~\cite{zolfaghari2018eco} &     	& 92  & 267$\times$1 & 150M & 46.4 & - & - & -     \\
ECO Two-stream~\cite{zolfaghari2018eco} &  \cmark   	& 92+92  & N/A$\times$1 & 300M & 49.5 & - & - & -     \\
I3D from~\cite{wang2018videos}& 	& 32 & 153G$\times$2 & 28.0M & 41.6 & 72.2 & - & - \\
NL-I3D from~\cite{wang2018videos}& 	& 32  & 168G$\times$2 & 35.3M & 44.4 & 76.0 & - & - \\
NL-I3D+GCN~\cite{wang2018videos}  &	& 32  & 303G$\times$2 & 62.2M & 46.1 & 76.8 & - & - \\
S3D-G~\cite{xie2018rethinking}    &	& 64 & 71G$\times$1 & 11.6M & 48.2 &   78.7 & - & -     \\
DFB-Net \cite{martinez2019action} & & 16 & N/A$\times$1 & - & 50.1 & 79.5 & - & -     \\
STM \cite{jiang2019stm} &	& 16   & 67G$\times$30 & 24.0M & 50.7 &   80.4 & 64.2 & 89.8    \\
\midrule
TSM~\cite{lin2019tsm} & 	& 8    & 33G$\times$1 & 24.3M & 45.6 &   74.2 & 58.8 & 85.4     \\
TSM~\cite{lin2019tsm} & 	& 16    & 65G$\times$1 & 24.3M & 47.3 &   77.1 & 61.2 & 86.9   \\
TSM$_{En}$~\cite{lin2019tsm} & & 16+8 & 98G$\times$1 & 48.6M & 49.7 &   78.5 & 62.9 & 88.1    \\
TSM Two-stream~\cite{lin2019tsm} & \cmark   	& 16+16  & 129G$\times$1 & 48.6M & 52.6 &   81.9 & 65.0$^{\dagger}$ & 89.4$^{\dagger}$\\
TSM Two-stream~\cite{lin2019tsm} & \cmark   	& 16+16  & 129G$\times$6 & 48.6M & - &   - &66.0 & 90.5\\
\midrule
MSNet-R50 (ours) &	& 8 & 34G$\times$1 & 24.6M & 50.9 & 80.3 & 63.0 & 88.4 \\ 
MSNet-R50 (ours) &	& 16 & 67G$\times$1 & 24.6M & 52.1 & 82.3 & 64.7 & 89.4  \\   
MSNet-R50$_{En}$ (ours) &	 & 16+8 & 101G$\times$1 & 49.2M & 54.4 & 83.8 & 66.6 & 90.6 \\ 
MSNet-R50$_{En}$ (ours) &	 & 16+8 & 101G$\times$10 & 49.2M & \textbf{55.1} & \textbf{84.0} & \textbf{67.1} & \textbf{91.0}  \\
\bottomrule
\end{tabular}
}
\end{table}

\subsection{Comparison with state-of-the-art methods}

Table~\ref{sota_table} summarizes the results on Something-Something V1\&V2.
Each section of the table contains results of 2D CNN methods~\cite{lee2018motion,liu2019learning,wang2016temporal,zhou2018temporal}, 3D CNN methods~\cite{jiang2019stm,martinez2019action,wang2018videos,xie2018rethinking,zolfaghari2018eco}, ResNet with TSM (TSM ResNet)~\cite{lin2019tsm}, and the proposed method, respectively. Most of the results are copied from the corresponding papers, except for TSM ResNet; we evaluate the official pre-trained model of TSM ResNet using a single center-cropped clip per video in terms of top-1 and top-5 accuracies.
Our method, which uses TSM ResNet as a backbone, achieves 50.9\% and 63.0\% on Something-Something V1 and V2 at top-1 accuracy, respectively, which outperforms most of 2D CNN and 3D CNN methods, while using a single clip with 8 input frames only.
Compared to the TSM ResNet baseline, our method obtains a significant gain of about 5.3\% points and 4.2\% points at top-1 accuracy at the cost of only 2.5\% and 1.2\% increase in FLOPs and parameters, respectively.
When using 16 frames, our method further improves achieving 52.1\% and 64.7\% at top-1 accuracy, respectively.
Following the evaluation procedure of two-stream networks, we evaluate the ensemble model (MSNet-R50$_{En}$) by averaging prediction scores of the 8-frame and 16-frame models.
With the same number of clips for evaluation, it achieves top-1 accuracy 1.8\% points and 1.6\% points higher than TSM two-stream networks with 22\% less computation, even no optical flow needed.
Our 10-clip model achieves 55.1\% and 67.1\% at top-1 accuracy on Something-Something V1 and V2, respectively, which is the state-of-the-art on both of the datasets.
As shown in Figure~\ref{fig:1}, our model provides the best trade-off in terms of accuracy, FLOPs, and the number of parameters.

\begin{table}[h]
\centering
\captionsetup{width=0.91\textwidth}
\caption{Performance comparison with motion representation methods.
The symbol ${\ddagger}$ denotes that we only report the backbone FLOPs.
\label{kinetics_table}}
\renewcommand{\arraystretch}{0.9}
\fontsize{8pt}{9.5pt}\selectfont
\scalebox{1.0}{
\begin{tabular}{lcccccc}
\toprule
model & flow & \#frame & FLOPs & speed & Kinetics & HMDB51 \\
 & & & $\times$clips & (V/s) & Top-1 & Top-1 \\
\midrule
ResNet-50 from~\cite{piergiovanni2019representation} &  & 32  & 132G$\times$25  & 22.8  & 61.3 & 59.4 \\
R(2+1)D~\cite{tran2018closer} &      & 32  & 152G$\times$115  & 8.7      & 72.0 & 74.3 \\
\midrule
MFNet from~\cite{piergiovanni2019representation} & & 10 & 80G$^{\ddagger}\times$10 & - & - & 56.8  \\
OFF(RGB)~\cite{sun2018optical} &  & 1 & N/A$\times$25 & - & - & 57.1  \\
TVNet~\cite{fan2018end} &  & 18 & N/A$\times$250 & - & - &71.0   \\
STM~\cite{jiang2019stm}   &    & 16  & 67G$\times$30  & -      & 73.7 & 72.2 \\
Rep-flow (ResNet-50)~\cite{piergiovanni2019representation} & & 32 & 132G$^{\ddagger}\times$25 & 3.7 & 68.5 & 76.4 \\
Rep-flow (R(2+1)D)~\cite{piergiovanni2019representation} & & 32 & 152G$^{\ddagger}\times$25 & 2.0 & 75.5 & 77.1  \\
\midrule
ResNet-50 Two-stream from~\cite{piergiovanni2019representation}   & \cmark & 32+32  & 264G$\times$25     & 0.2    & 64.5 & 66.6  \\
R(2+1)D Two-stream~\cite{tran2018closer}   & \cmark    & 32+32  & 304G$\times$115    & 0.2     & 73.9 & \textbf{78.7}  \\
OFF(RGB+Flow+RGB Diff)~\cite{sun2018optical} & \cmark & 1+5+5 & N/A$\times$25  & - & - & 74.2  \\
\midrule
TSM (reproduced)    &  & 8  & 33G$\times$10     & 64.1       & 73.5  & 71.9\\
MSNet-R50 (ours)	    &                    &  8 & 34G$\times$10  & 54.2  & 75.0 & 75.8 \\ 
MSNet-R50 (ours)		&                        &  16 & 67G$\times$10  & 31.2  & \textbf{76.4} &77.4 \\ 
	\bottomrule
\end{tabular}
}
\end{table}

\subsection{Comparison with other motion representation methods}
Table~\ref{kinetics_table} summarizes comparative results with other motion representation methods~\cite{lee2018motion,sun2018optical,fan2018end,piergiovanni2019representation,jiang2019stm} based on RGB frames.
The comparison is done on Kinetics and HMDB51 since the previous methods commonly report their results on them.
Each section of the table contains results of conventional 2D and 3D CNNs, motion representation methods~\cite{lee2018motion,sun2018optical,fan2018end,piergiovanni2019representation,jiang2019stm}, two-stream CNNs with optical flows~\cite{he2016deep,tran2018closer}, and the proposed method, respectively.
OFF, MFNet, and STM~\cite{lee2018motion,sun2018optical,jiang2019stm} use a sub-network or lightweight modules to calculate temporal gradients of frame-wise feature maps.
TVNet~\cite{fan2018end} and Rep-flow~\cite{piergiovanni2019representation} internalize iterative TV-L1 flow operations in their networks.
As shown in Table~\ref{kinetics_table}, the proposed model using 16 frames outperforms all the other conventional CNNs and the motion representation methods~\cite{he2016deep,carreira2017quo,tran2018closer,lee2018motion,sun2018optical,fan2018end,piergiovanni2019representation,jiang2019stm}, while being competitive with the R(2+1)D two-stream~\cite{tran2018closer} that uses pre-computed optical flows.
Furthermore, our model is highly efficient than all the other methods in terms of FLOPs, clips, and the number of frames.

\smallbreak
\noindent
\textbf{Run-time.}
 We also evaluate in Table \ref{kinetics_table} the inference speeds of several models to demonstrate the efficiency of our method.
All the run-times reported are measured on a single GTX Titan Xp GPU, ignoring the time of data loading. For this experiment, 
we set the spatial size of the input to $224\times 224$ and the batch size to 1. 
The official codes are used for ResNet, TSM ResNet, and Rep-flow~\cite{he2016deep,lin2019tsm,piergiovanni2019representation} except for R(2+1)D\cite{tran2018closer} we implemented.
In evaluating Rep-flow~\cite{piergiovanni2019representation}, we use 20 iterations for optimization as in the original paper.
The speed of the two-stream networks~\cite{simonyan2014two,tran2018closer} includes computation time for TV-L1 method on the GPU.
The run-time results clearly show the cost of iterative optimizations used in two-stream networks and Rep-flow. 
In contrast, our model using 16 frames is about 160$\times$ faster than the two-stream networks.
Compared to Rep-flow ResNet-50, our method performs about 4$\times$ faster due to the absence of the iterative optimization process in Rep-flow.

\subsection{Ablation studies} \label{ablation}
We conduct ablation studies of the proposed method on Something-Something V1~\cite{goyal2017something} dataset.
We use ImageNet pre-trained TSM ResNet-18 as a default backbone and use 8 input frames for all experiments in this section.

\smallbreak
\noindent
\textbf{Displacement estimation in MS module.}  
In Table~\ref{Displacement_table}, we experiment with different variants of the displacement tensor $\mathbf{D}^{(t)}$ in the MS module. 
We first compare soft-argmax (`S') and kernel-soft-argmax (`KS') for displacement estimation. As shown in the upper part of Table~\ref{Displacement_table}.
the kernel-soft-argmax outperforms the soft-argmax, showing the noise reduction effect of Gaussian kernel.
In the lower part of Table~\ref{Displacement_table}, we evaluate the effect of additional features: confidence maps (`CM') and backward displacement tensor (`BD').
The backward displacement tensor is estimated from $\textbf{F}^{(t+1)}$ to $\textbf{F}^{(t)}$.
We concatenate the forward and backward displacement tensors, and then pass them to the feature transformation layers.
We obtain 0.9\% points gain by appending the confidence map to the displacement tensor.
Furthermore, by adding backward displacement we obtain another 0.5\% points gain at top-1 accuracy, indicating that forward and backward displacement maps complement each other to enrich motion information.
We use the kernel-soft-argmax with the confidence map (`KS + CM') as a default method for all other experiments.

\begin{table}[t]
    \begin{minipage}{1.0\textwidth}
        \begin{minipage}{0.49\textwidth}
            \fontsize{8.5pt}{9.5pt}\selectfont
            \captionsetup{width=\columnwidth}
            \caption{Performance comparison with different displacement estimations.} 
            \begin{tabular}[t]{L{2.2cm}|C{1.1cm}|C{1.1cm}|C{1.1cm}} 	
                \toprule
                model           & FLOPs     & top-1 & top-5 \\
                \hline
                baseline        & 14.6G     & 41.5 & 71.8  \\
                 \hline
                S               & 14.8G     & 43.8 & 74.9 \\   
                KS              & 14.9G     & 44.6 & 75.4 \\ 
                \hline
                KS + CM         & 14.9G     & 45.5 & 76.5 \\
                KS + CM + BD    & 15.1G     & 46.0 & 76.7 \\
                \bottomrule
            \end{tabular}
            \label{Displacement_table}
        \end{minipage}
        \hfill
        \begin{minipage}{0.47\textwidth}
            \includegraphics[width=\columnwidth]{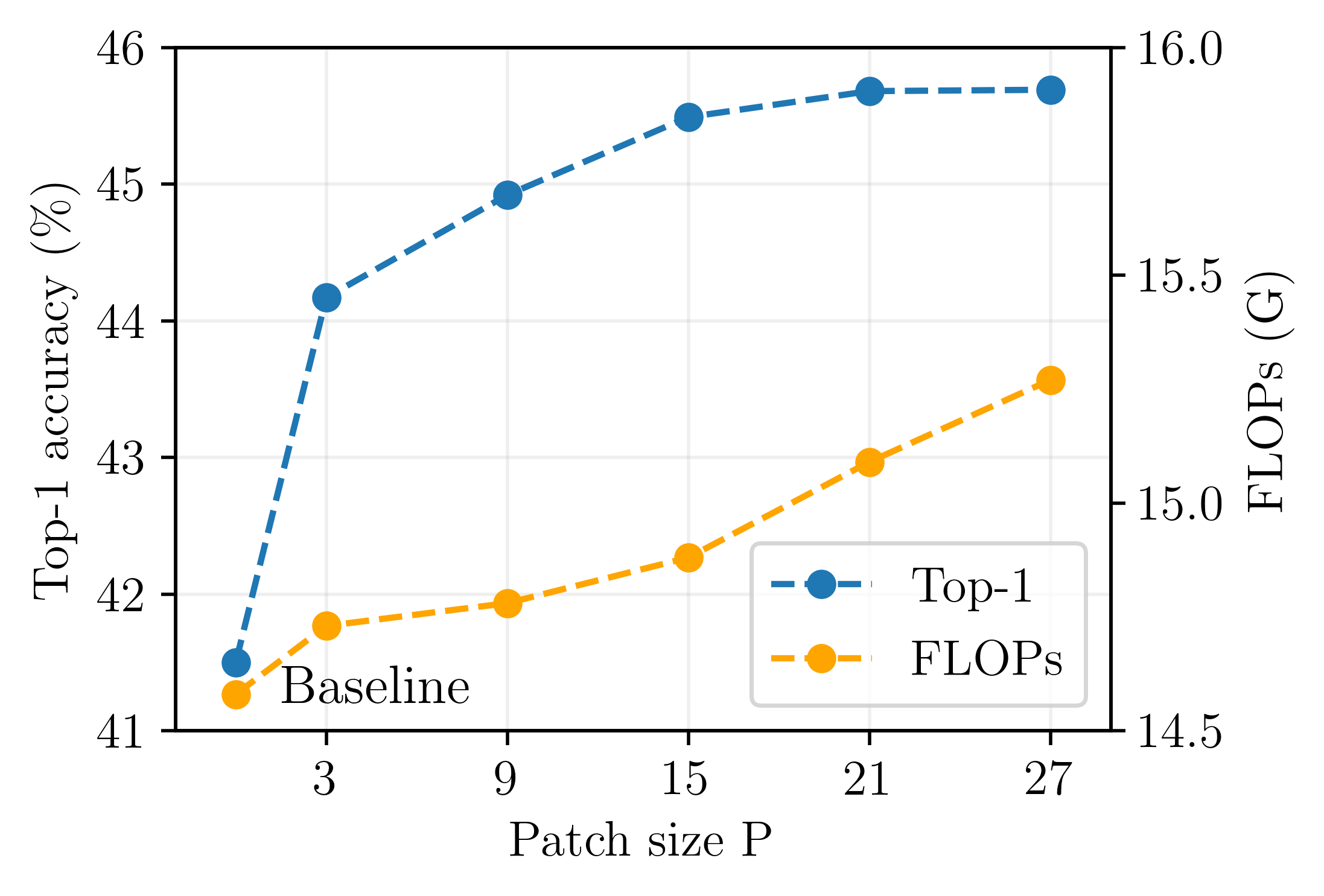}
            \captionsetup{width=1.0\columnwidth}
            \captionof{figure}{Top-1 accuracy and FLOPs with different patch sizes.}
        \label{fig:4}
        \end{minipage}
        
        \begin{minipage}{0.48\textwidth}
            \fontsize{8.5pt}{9.5pt}\selectfont
            \captionsetup{width=1.0\columnwidth}
            \caption{Performance comparison with different positions of the MS module.} 
            \begin{tabular}[t]{L{2.2cm}|C{1.1cm}|C{1.1cm}|C{1.1cm}} 
            \toprule
            model & FLOPs & top-1 & top-5 \\
            \hline
            baseline            & 14.6G      & 41.5 & 71.8     \\
            \hline
            $res_2$             & 15.6G     & 45.1 & 76.1     \\
            $res_3$             & 14.9G     & 45.5 & 76.5    \\
            $res_4$             & 14.7G     & 42.6 & 73.2   \\
            $res_5$             & 14.6G     & 41.1  & 71.8    \\
            $res_{2,3,4}$       & 16.0G     & 45.7  & 76.8  \\
            \bottomrule
            \end{tabular}
            \label{position_table}
        \end{minipage}
        \hfill
        \begin{minipage}{0.48\textwidth}
            \captionsetup{width=0.95\columnwidth}
            \caption{Performance comparison with different fusing strategies.}
            \fontsize{8.5pt}{9.5pt}\selectfont
            \begin{tabular}[t]{L{2.2cm}|C{1.1cm}|C{1.1cm}|C{1.1cm}} 
            \toprule
            model & FLOPs & top-1 & top-5 \\
            \hline
            baseline            & 14.6G     & 41.5 & 71.8     \\
            \hline
            MS only             & 14.1G     & 38.8 & 70.7     \\
            multiply            & 14.9G     & 44.5 & 75.9   \\
            concat.             & 15.7G     & 45.0 & 76.1    \\      
            add                 & 14.9G     & 45.5  & 76.5    \\
            \bottomrule
            \end{tabular}
            \label{fusion_table}
        \end{minipage}
    \end{minipage}
\end{table}

\noindent
\textbf{Size of matching region.}
In Figure~\ref{fig:4}, we evaluate performance varying the spatial size of matching regions of the MS module.
Even with a small matching region $P=3$, it provides a noticeable performance gain of over $2.7\%$ points to the baseline.
The performance tends to increase as the matching region becomes larger due to the larger displacement it can handle between frames.
The performance is saturated after $P=15$. 

\smallbreak
\noindent \textbf{Position of MS module.}
In Table~\ref{position_table}, we evaluate different positions of the MS module.
We denote that $res_{N}$ by the $N$-th stage of the ResNet.
For each stage, it is inserted right after its final residual block. 
The result shows that while the MS module is beneficial in most cases, both accuracy and efficiency gains depend on the position of the module.
It performs the best at $res_3$; appearance features from $res_2$ are not strong enough for accurate feature matching while spatial resolutions of appearance features from $res_4$ and $res_5$ are not high enough.
The position of the module also affects FLOPs; the computational cost quadratically increases with spatial resolution due to convolution layers of the feature transformation.
When inserting multiple MS modules ($res_{2,3,4}$) at the backbone, it marginally improves top-1 accuracy as 0.2\% points.
Multiple modules appear to generate similar motion information even in different levels of features.

\smallbreak
\noindent \textbf{Fusing strategy of MS module.}
In Table \ref{fusion_table}, we evaluate different fusion strategies for the MS module; `MS only', `multiply', `concat', and `add'.
In the case of `MS only', we only pass $\mathbf{M}^{(t)}$ into downstream layers without $\mathbf{F}^{(t)}$.
We apply element-wise multiplication and element-wise addition, respectively, for `multiply' and `add'.
In the case of `concat', we concatenate $\mathbf{F}^{(t)}$ and $\mathbf{M}^{(t)}$, whose channel size is transformed to $C$ via an $1\times1\times1$ convolution layer.
`MS only' is less accurate than the baseline because visual semantic information is discarded.
While both `multiply' and `concat' clearly improve the accuracy, `add' achieves the best performance with 45.5\% at top-1 accuracy. 
We find that additive fusion is the most effective and stable in amplifying appearance features of moving objects. 

\begin{table}[t]
\begin{minipage}{1.0\textwidth}
    \begin{minipage}{0.41\textwidth}
        \centering
        \includegraphics[width=\columnwidth]{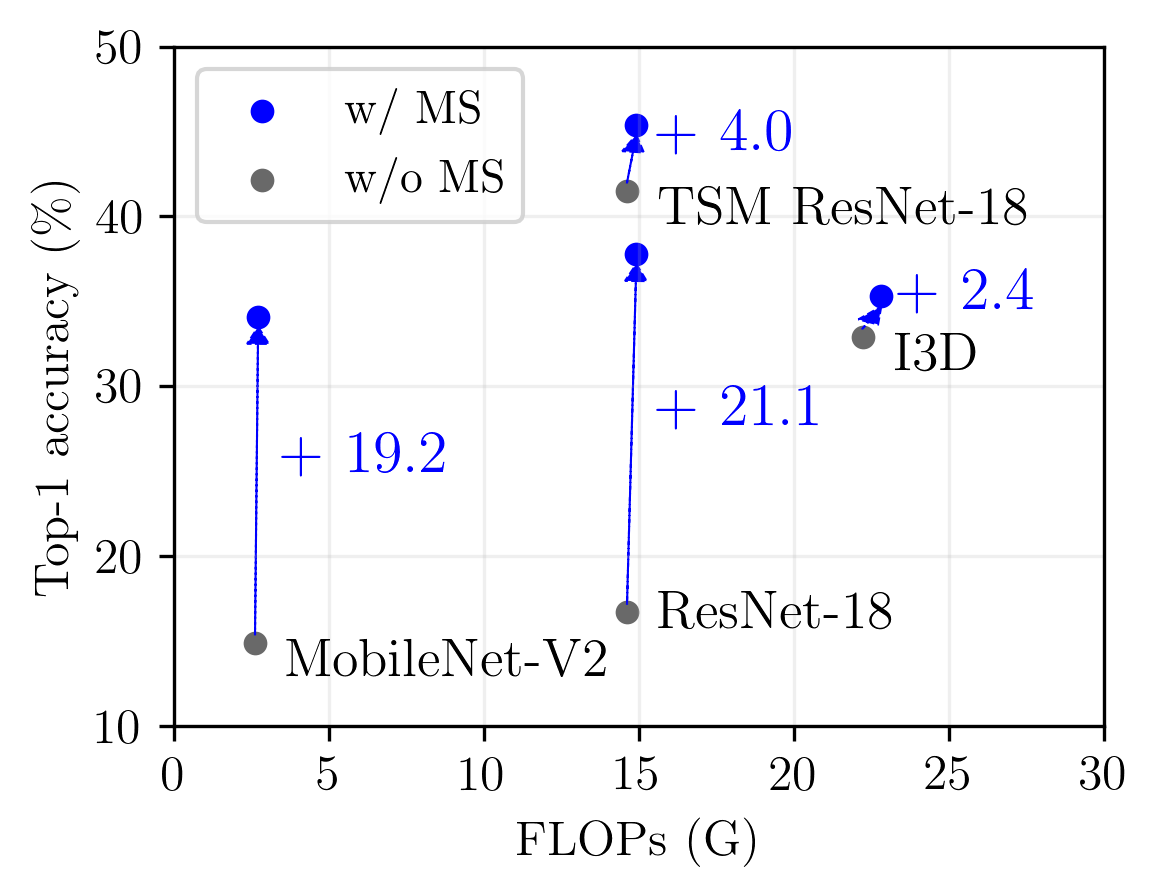}
        \captionsetup{width=0.95\columnwidth}
        \captionof{figure}{Top-1 accuracy and FLOPs with MS module on different backbones.}
        \label{fig:5}
    \end{minipage}
    \hfill
    \begin{minipage}{0.58\textwidth}
    \centering
    \fontsize{8pt}{9.5pt}\selectfont
    \captionof{table}{Performance comparison with two-stream networks.}
    \begin{tabular}[t]{l|C{0.8cm}|c|C{0.8cm}|C{0.8cm}} 
    \toprule
      model    &  flow &   FLOPs & top-1   &   top-5   \\
      \hline
      baseline         &  & 14.6G  & 41.5 & 71.8     \\
      \hline
      Two-stream$_{8+(8\times5)}$   & \cmark & 31.4G  & 46.8 & 77.3     \\
      Two-stream$_{8+(8\times1)}$     & \cmark & 28.9G & 44.7 & 75.2     \\
      Two-stream$_{8+(8\times1) (low)}$      & \cmark & 28.9G & 44.1 & 74.9     \\     
      \hline
      MSNet  &  & 14.9G   & 45.5 & 76.5     \\
      \bottomrule
    \end{tabular}
    \label{twostream_table}
    \end{minipage}
\end{minipage}
\end{table}

\smallbreak
\noindent 
\textbf{Effect of MS module on different backbones.}
In Figure~\ref{fig:5}, we also evaluate the effect of the MS module on ResNet-18, MobileNet-V2, and I3D.
We insert one MS module where the spatial resolution of the feature map remains the same.
For ResNet-18 and MobileNet-V2, we finetune models pre-trained on ImageNet. We train I3D from scratch.
Our MS module benefits both 2D CNNs and 3D CNNs to obtain higher accuracy.
The module significantly improves ResNet-18 and MobileNet-V2 by 21.3\% and 19.2\% points, respectively, in top-1 accuracy.
Since 2D CNNs do not use any spatio-temporal features, it obtains significantly higher gain from the MS module.
The MS module also improves I3D and TSM ResNet-18 by 2.4\% and 4.0\% points, respectively, in top-1 accuracy.
The gain on 3D CNNs, although relatively small, verifies that the motion features by the MS module are complementary even to the spatio-temporal features; the MS module learns explicit motion information across adjacent frames whereas TSM covers long-term temporal length using (pseudo-)temporal convolutions.

\smallbreak
\noindent
\textbf{Comparison with two-stream networks.}
In Table~\ref{twostream_table}, we compare the proposed method with variants of TSM two-stream networks~\cite{simonyan2014two} that use TV-L1 optical flows~\cite{zach2007duality}.
We denote the two-stream networks by Two-stream$_{N_r+(N_f \times N_s)}$ where $N_r$, $N_f$ and $N_s$ indicate the number of frames, optical flows, and their stacking size, respectively.
For each frame, the two-stream networks use $N_s$ stacked optical flows, which are extracted using the subsequent frames in the original video.
Note that those frames for optical flow extraction are not used in our method (MSNet). 
The second row of Table~\ref{twostream_table}, Two-stream$_{8+(8\times5)}$, shows the performance of standard TSM two-stream networks that use 5 stacked optical flows for the temporal stream.
Using the multiple optical flows for each frame outperforms our model in terms of accuracy but requires substantially larger FLOPs as well as an additional computation for calculating optical flows. 
For a fair comparison, we report the performance of the two-stream networks, Two-stream$_{8+(8\times1)}$, that do not stack multiple optical flows.
Our model outperforms the two-stream networks by 0.8\% points at top-1 accuracy, with about two times fewer FLOPs.
Note that both Two-stream$_{8+(8\times5)}$ and Two-stream$_{8+(8\times1)}$ use optical flows obtained from the original video with a higher frame rate than the input video clip (sampled frames); our method (MSNet) observes the input video clip only.  
We thus evaluate other two-stream networks, Two-stream$_{8+(8\times1) (low)}$, that uses low-fps optical flows as input; we sample a sequence of frames in 3 fps from the original video and extract TV-L1 optical flows using the sequence. 
As shown in Table~\ref{twostream_table}, the top-1 accuracy gap between ours and the two-stream network increases to 1.4\% points. The result implies that given low-fps videos, our method may further improve over the two-stream networks.

\subsection{Visualization}
\begin{figure}[t]
    \centering
    \begin{subfigure}[]{0.495\columnwidth}
    \includegraphics[width=\columnwidth]{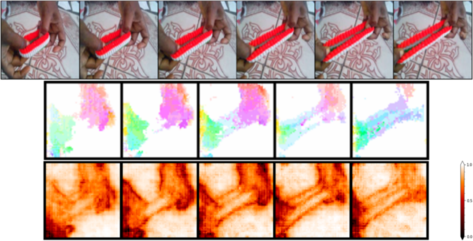}
    \caption{Label: ``Pulling two ends of something so that it gets stretched."}
    \label{fig:6a}
    \end{subfigure}
    \begin{subfigure}[]{0.495\columnwidth}
    \includegraphics[width=\columnwidth]{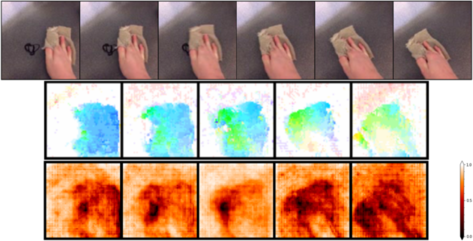}
    \caption{Label:    ``Wiping off something of something."}
    \label{fig:6b}
    \end{subfigure}
    \begin{subfigure}[]{0.495\columnwidth}
    \includegraphics[width=\columnwidth]{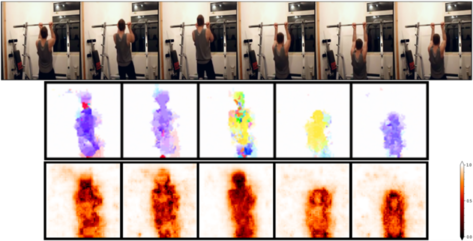}
    \caption{Label: ``Pull ups."}
    \label{fig:6c}
    \end{subfigure}
    \begin{subfigure}[]{0.495\columnwidth}
    \includegraphics[width=\columnwidth]{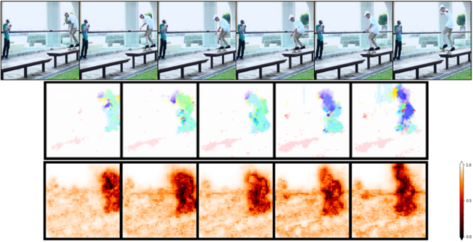}
    \caption{Label: ``Skateboarding."}
    \label{fig:6d}
    \end{subfigure}
\caption{Visualization on Something-Something-V1 (top) and Kinetics (bottom) datasets. RGB images, displacement maps, and the confidence maps are shown from the top row in each subfigure.} \label{fig:6}
\end{figure}

In Figure~\ref{fig:6}, we present visualization results on Something-Something V1 and Kinetics datasets.
They show that our MS module effectively learns to estimate motion without any direct supervision used in training. 
The first row of each subfigure shows 6 uniformly sampled frames from a video.
The second and third rows show color-coded displacement maps~\cite{baker2011database} and confidence maps, respectively; we apply min-max normalization on the confidence map.
The resolution of all the displacement and confidence maps is set to 56$\times$56 for better visualization.
As shown in the figures, the MS module captures reliable displacements in most cases: horizontal and vertical movements (Figure~\ref{fig:6a},~\ref{fig:6c},~\ref{fig:6d}), rotational movements (Figure~\ref{fig:6b}), and non-severe deformation (Figure~\ref{fig:6a},~\ref{fig:6d}).
See the supplementary material for additional details and results. We will make our code and data available online.

\section{Conclusion}
We have presented an efficient yet effective motion feature block, the MS module, that learns to generate motion features on the fly for video understanding. The MS module can be readily inserted into any existing video architectures and trained by backpropagation.
The ablation studies on the module demonstrate the effectiveness of the proposed method in terms of accuracy, computational cost, and model size.
Our method outperforms existing state-of-the-art methods on Something-Something-V1\&V2 for video classification with only a small amount of additional cost.

\bigskip
\noindent \textbf{Acknowledgements.}
This work is supported by Samsung Advanced Institute of Technology (SAIT), and also by Basic Science Research Program (NRF-2017R1E1A1A010 77999, NRF-2018R1C1B6001223) and Next-Generation Information Computing Development Program (NRF-2017M3C4A7069369) through the National Research Foundation of Korea (NRF) funded by the Ministry of Science, ICT.

\bibliographystyle{splncs04}
\bibliography{cvlab_cho}







\clearpage

\newcommand{\Fig}[1]{Figure~\ref{fig:#1}}
\newcommand{\Sec}[1]{Section~\ref{sec:#1}}
\newcommand{\Eq}[1]{Equation~(\ref{eq:#1})}
\newcommand{\Tbl}[1]{Table~\ref{tab:#1}}

\def\httilde{\mbox{\tt\raisebox{-.5ex}{\symbol{126}}}}

\begin{center}
\textbf{\large Supplementary Material of \enquote{MotionSqueeze: Neural Motion Feature Learning for Video Understanding}}
\end{center}

\setcounter{section}{0}


We present additional results and details that are omitted in our main paper due to the lack of space. All our code and data are released online at our project page: \url{http://cvlab.postech.ac.kr/research/MotionSqueeze/} 

\section{Effects of depth-wise separable (DWS) convolutions}
We use DWS convolutions rather than standard convolutions to build the feature transformation (FT) layers deeper and wider while saving computational cost. Table~\ref{table:dws} shows the results of different forms of FT layers on Something-Something V1~\cite{goyal2017something} . The accuracy increases as the FT layers become deeper and have wider receptive fields, and the DWS convolutions show the best accuracy-FLOPs tradeoff.

\section{Comparison with the CP module~\cite{liu2019learning}}
As we mentioned in the main paper, the CP module is the one of the most relevant work to our method in the sense that it leverage correspondences between input video frames. Here we provide more detailed comparisons to it.

\smallbreak
\noindent {\bf Difference in motivation and design.} Unlike our MS module, which focuses on extracting effective motion features across consecutive frames, the CP module~\cite{liu2019learning} is designed to capture long-term spatio-temporal relationship within an input video~\cite{wang2018non} by computing a non-local correlation tensor across all frames.
The CP module selects $k$ most likely corresponding features in the correlation tensor with an `$\mathrm{arg}$ top-$k$' operation, and the operation thus makes the correlation tensor non-differentiable.

\smallbreak
\noindent {\bf Performance comparison.}
We have already shown in Table 1 of the main paper that the result of our method is better than that of the CP module (from the original paper~\cite{liu2019learning}) on Something-Something V2~\cite{goyal2017something}. The comparison, however, may not be totally fair in the sense that the backbone and the other experimental settings are not the same. For an apples-to-apples comparison between the MS module and the CP module, we conduct an additional experiment using the same backbone and setup. We re-implement the CP module in Pytorch based on the official Tensorflow code\footnote{\url{https://github.com/xingyul/cpnet}}. As a baseline network, we use ImageNet pre-trained TSM ResNet-18 using 8 input frames.
Either MS or CP module is inserted after the third stage of the network.
Table~\ref{table:cp} summarizes the comparative results of the MS module and the CP module on Something-Something V1~\cite{goyal2017something}.
The CP module is effective for improving accuracy while consuming almost 6G FLOPs more than the baseline; the computational cost of the non-local correlation tensor is quadratic to the number of input frames. In contrast, the MS module performs 0.9\% points and 0.8\% points higher at top-1 and top-5 accuracy, respectively, while consuming 26\% less FLOPs, compared to the CP module.

\begin{table}[t]
	\centering
	\caption{Performance comparison with different forms of feature transformation (FT) layers. $n\times(k,k)$ denotes $n$ standard convolution layers with a kernel size of $k$. * denotes our FT layers in Fig. 3 of the paper.
		}
	\label{table:dws}
    \begin{tabular}[t]{L{2.5cm}|C{2.0cm}|C{1.2cm}|C{1.2cm}}  
    \toprule
    model & FT layers & FLOPs & Top-1 \\
    \hline
      TSM-R50     & -     & 33.1G & 46.7    \\
      \hline
      MSNet-R50   & $1\times(1,1)$     & 33.4G & 49.3     \\
      MSNet-R50   & $1\times(3,3)$     & 33.5G & 49.8     \\      
      MSNet-R50   & $4\times(3,3)$     & 35.8G & 50.4     \\      
      MSNet-R50   & ours*     & 33.7G & \textbf{50.9}     \\            
      \bottomrule
    \end{tabular}
\end{table}

\begin{table}[t]
	\centering
	\caption{Performance comparison between the CP module~\cite{liu2019learning} and the MS module.
		}
	\label{table:cp}
    \begin{tabular}[t]{L{2.5cm}|C{1.2cm}|C{1.2cm}|C{1.2cm}}  
    \toprule
    model & FLOPs & Top-1 & Top-5 \\
    \hline
      baseline     & 14.6G     & 41.5 & 71.8    \\
      \hline
      CP module~\cite{liu2019learning}   & 20.4G     & 44.9 & 75.6     \\
      MS module  & 15.0G     & 45.8 & 76.4      \\
      \bottomrule
    \end{tabular}
\end{table}

\section{Backbone architectures in experiments}
In our main paper, we evaluate the effect of the MS module on different backbone architectures: ResNet~\cite{he2016deep}, TSM ResNet~\cite{lin2019tsm}, MobileNet-V2~\cite{sandler2018mobilenetv2} and I3D~\cite{carreira2017quo}.
We provide details of the backbone architectures here.

\smallbreak
\noindent \textbf{ResNet \& TSM ResNet.}
Table~\ref{tab:resnet} shows the architecture of ResNet~\cite{he2016deep} and TSM ResNet~\cite{lin2019tsm}.
As a default, one MS module is inserted right after $res_3$.

\begin{figure}[t]
    \centering
    \begin{subfigure}[t]{0.4\columnwidth}
    \centering    
    \includegraphics[width=\columnwidth]{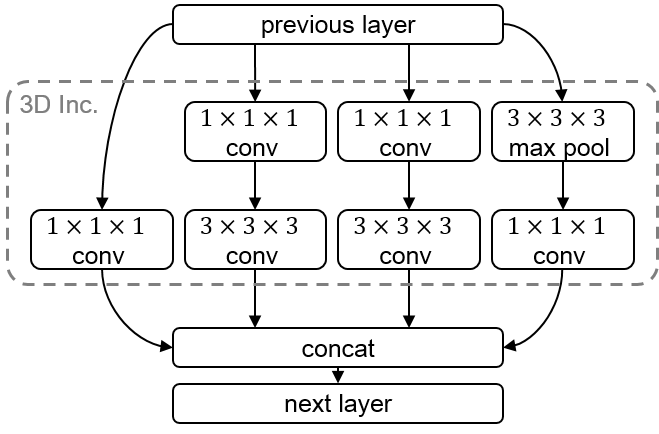}
    \caption{A 3D Inception module of I3D~\cite{carreira2017quo}}
    \label{fig:s2a}
    \end{subfigure}
    \qquad
    \centering
    \begin{subfigure}[t]{0.52\columnwidth}
    \centering    
    \includegraphics[width=\columnwidth]{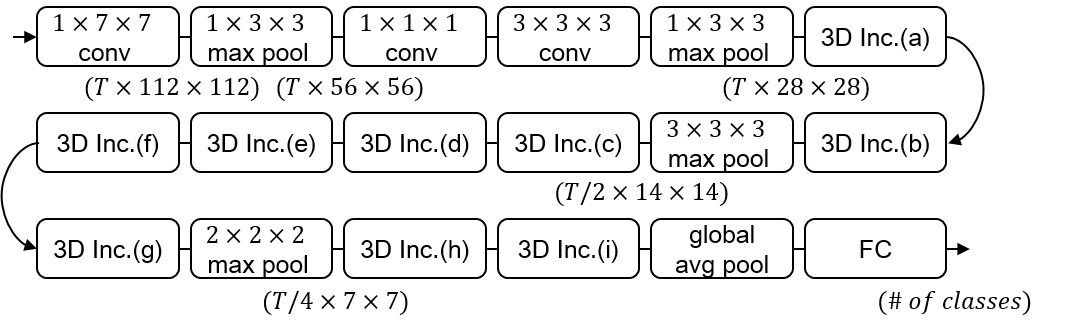}
    \caption{I3D~\cite{carreira2017quo} architecture.}
    \label{fig:s2b}
    \end{subfigure}
\caption{I3D (BN-Inception~\cite{ioffe2015batch}) backbone.} \label{fig:s2}
\end{figure}

\smallbreak
\noindent \textbf{I3D.}
Figure~\ref{fig:s2a},~\ref{fig:s2b} show the architecture of I3D~\cite{carreira2017quo} used in our experiment; we reduce the first convolution kernel from 7$\times$7$\times$7 to 1$\times$7$\times$7 as we only use a sampled clip of $8$ frames.
The MS module is inserted after $Inc(b)$ of Figure~\ref{fig:s2b}.

\smallbreak
\noindent \textbf{MobileNet-V2.}
Figure~\ref{fig:s1} and Table~\ref{tab:mobile} show the architecture of MobileNet-V2~\cite{sandler2018mobilenetv2}.
The MS module is inserted right after $stage_3$ of Table~\ref{tab:mobile}.
As the feature channel size of the backbone is small enough, we omit the channel reduction layer in the MS module. 

\section{Additional examples of visualization}
\begin{figure*}[t]
    \centering
    \begin{subfigure}[t]{0.48\columnwidth}
    \includegraphics[width=\columnwidth]{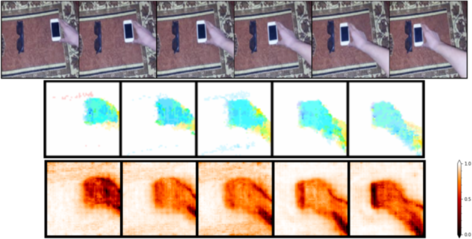}
    \caption{``Moving sth. closer to sth.".}
    \label{fig:s3a}
    \end{subfigure}
    \begin{subfigure}[t]{0.48\columnwidth}
    \includegraphics[width=\columnwidth]{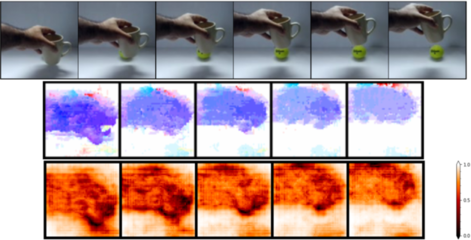}
    \caption{``Removing sth., revealing sth. behind".}
    \label{fig:s3b}
    \end{subfigure}
    \begin{subfigure}[t]{0.48\columnwidth}
    \includegraphics[width=\columnwidth]{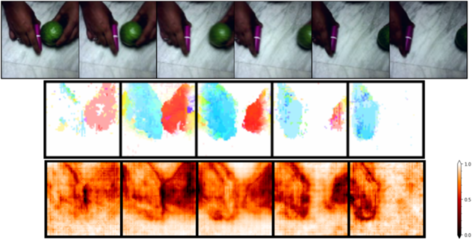}
    \caption{``Moving sth. and sth. away from each other".}
    \label{fig:s3c}
    \end{subfigure}
    \begin{subfigure}[t]{0.48\columnwidth}
    \includegraphics[width=\columnwidth]{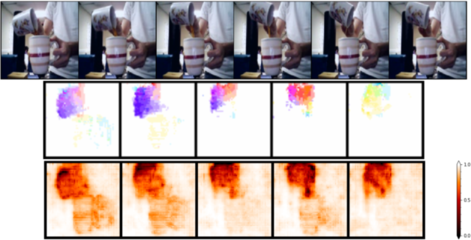}
    \caption{``Pouring sth. into sth.".}
    \label{fig:s3d}
    \end{subfigure}
    \begin{subfigure}[t]{0.48\columnwidth}
    \includegraphics[width=\columnwidth]{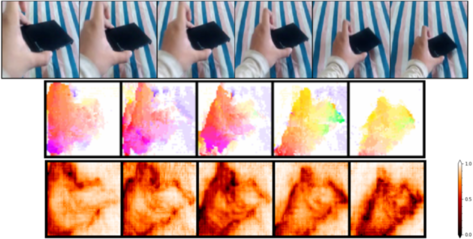}
    \caption{``Pretending to put sth. on a surface".}
    \label{fig:s3e}
    \end{subfigure}
    \begin{subfigure}[t]{0.48\columnwidth}
    \includegraphics[width=\columnwidth]{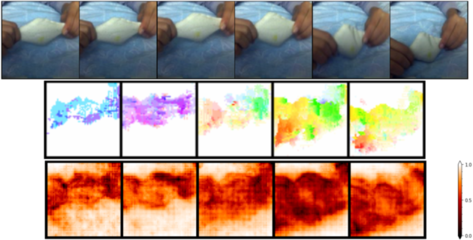}
    \caption{``Pulling two ends of sth. so that it gets stretched".}
    \label{fig:s3f}
    \end{subfigure}
    \begin{subfigure}[t]{0.48\columnwidth}
    \includegraphics[width=\columnwidth]{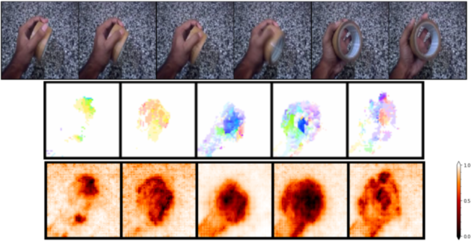}
    \caption{``Pretending to squeeze sth.".}
    \label{fig:s3g}
    \end{subfigure}
    \begin{subfigure}[t]{0.48\columnwidth}
    \includegraphics[width=\columnwidth]{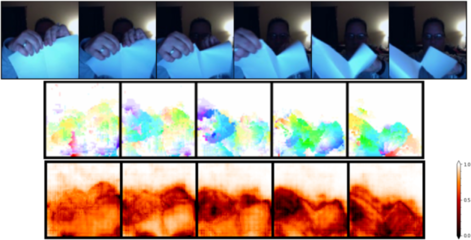}
    \caption{``Tearing sth. into two pieces".}
    \label{fig:s3h}
    \end{subfigure}
\caption{Visualization on Something-Something V1~\cite{goyal2017something} dataset. Video frames, displacement maps, and confidence maps are shown from the top row in each subfigure.} \label{fig:s3}
\end{figure*}

\begin{figure*}[t]
    \centering
    \begin{subfigure}[t]{0.49\columnwidth}
    \includegraphics[width=\columnwidth]{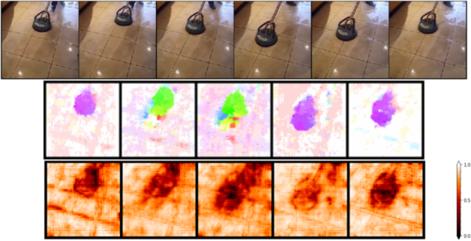}
    \caption{``Cleaning floor".}
    \label{fig:s4a}
    \end{subfigure}
    \begin{subfigure}[t]{0.49\columnwidth}
    \includegraphics[width=\columnwidth]{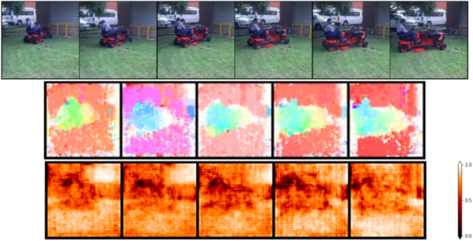}
    \caption{``Cleaning pool".}
    \label{fig:s4b}
    \end{subfigure}
    \begin{subfigure}[t]{0.49\columnwidth}
    \includegraphics[width=\columnwidth]{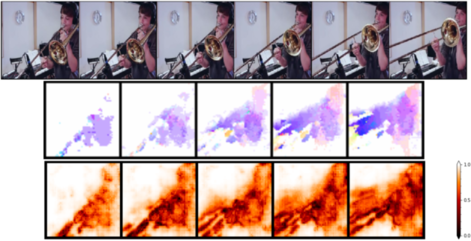}
    \caption{``Playing trombone".}
    \label{fig:s4c}
    \end{subfigure}
    \begin{subfigure}[t]{0.49\columnwidth}
    \includegraphics[width=\columnwidth]{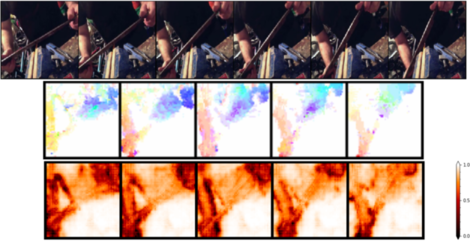}
    \caption{``Sharpening pencil".}
    \label{fig:s4d}
    \end{subfigure}
    \begin{subfigure}[t]{0.49\columnwidth}
    \includegraphics[width=\columnwidth]{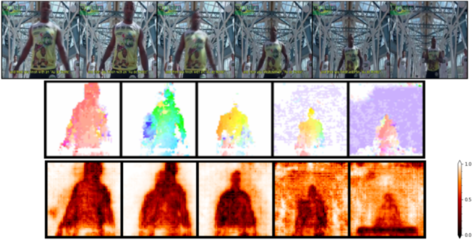}
    \caption{``Capoeira".}
    \label{fig:s4e}
    \end{subfigure}
    \begin{subfigure}[t]{0.49\columnwidth}
    \includegraphics[width=\columnwidth]{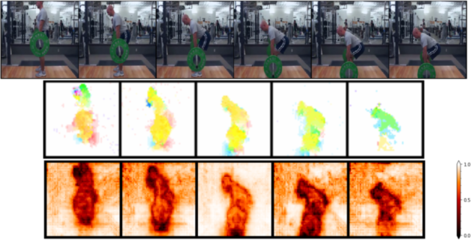}
    \caption{``Deadlifting".}
    \label{fig:s4f}
    \end{subfigure}
    \begin{subfigure}[t]{0.49\columnwidth}
    \includegraphics[width=\columnwidth]{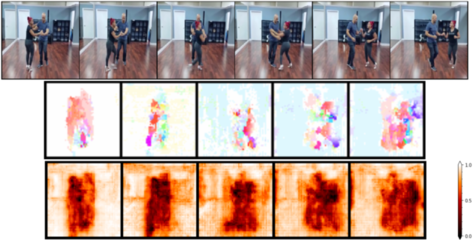}
    \caption{``Salsa dancing".}
    \label{fig:s4g}
    \end{subfigure}
    \begin{subfigure}[t]{0.49\columnwidth}
    \includegraphics[width=\columnwidth]{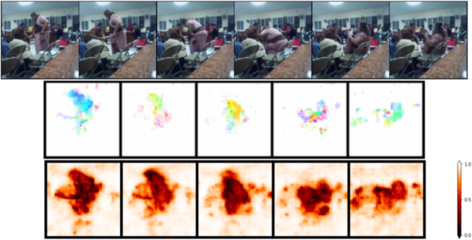}
    \caption{``Washing feet".}
    \label{fig:s4h}
    \end{subfigure}
\caption{Visualization on Kinetics-400~\cite{kay2017kinetics} dataset. Video frames, displacement maps, and confidence maps are shown from the top row in each subfigure.} \label{fig:s4}
\end{figure*}

We present more results of visualization on Something-Something V1~\cite{goyal2017something} in Figure~\ref{fig:s3} and Kinetics-400~\cite{kay2017kinetics} in Figure~\ref{fig:s4}.
From the top of each figure, RGB frames, color-coded displacement maps~\cite{baker2011database}, and confidence maps are illustrated.
We visualize examples of horizontal, vertical movements (Figure~\ref{fig:s3a},~\ref{fig:s3b},~\ref{fig:s3c},~\ref{fig:s4a},~\ref{fig:s4b},~\ref{fig:s4c}), rotations (Figure~\ref{fig:s3d},~\ref{fig:s4d}), scale changes (Figure~\ref{fig:s3e},~\ref{fig:s4e}), and deformations (Figure~\ref{fig:s3f},~\ref{fig:s4f}).
We also report some failure cases in the last row of figures (Figure~\ref{fig:s3g},~\ref{fig:s3h},~\ref{fig:s4g},~\ref{fig:s4h}); estimated displacement maps around regions of occlusion or severe deformation are often inaccurate.   

\clearpage

\begin{table}[t]
  \begin{center}
    \caption{ResNet \& TSM ResNet backbone.}
    \label{tab:resnet}   
    \scalebox{0.8}{
    \begin{tabular}{c|c|c|c|c|c}
    \hline
      Layers & ResNet-18 & TSM ResNet-18 & ResNet-50 & TSM ResNet-50 &Output size \\
 \hline 
      conv$_1$ & \multicolumn{4}{c|}{1$\times$7$\times$7, 64, stride 1,2,2} & T$\times$112$\times$112 \\
  \hline
      \multirow{4}{*}{res$_2$}& \multicolumn{4}{c|}{1$\times$3$\times$3 max pool, stride 2}  & \multirow{4}{*}{T$\times$56$\times$56 } \\  
      \cline{2-5}  
      & $\begin{bmatrix}  $1$\times$3$\times$3, 64$ \\    $1$\times$3$\times$3, 64$ \end{bmatrix}\times$2  
      & $\begin{bmatrix}  $TSM$ \\$1$\times$3$\times$3, 64$ \\    $1$\times$3$\times$3, 64$ \end{bmatrix}\times$2  
      & $\begin{bmatrix}  $1$\times$1$\times$1, 256$ \\ $1$\times$3$\times$3, 256$ \\    $1$\times$1$\times$1, 256$ \end{bmatrix}\times$3  
      & $\begin{bmatrix}  $TSM$ \\ $1$\times$1$\times$1, 256$ \\ $1$\times$3$\times$3, 256$ \\    $1$\times$1$\times$1, 256$ \end{bmatrix}\times$3 
      \\ 
    \hline
  res$_3$ & $\begin{bmatrix} $1$\times$3$\times$3, 128$ \\    $1$\times$3$\times$3, 128$ \end{bmatrix}\times$2 
  & $\begin{bmatrix} $TSM$\\$1$\times$3$\times$3, 128$ \\    $1$\times$3$\times$3, 128$ \end{bmatrix}\times$2 
  & $\begin{bmatrix} $1$\times$1$\times$1, 512$ \\  $1$\times$3$\times$3, 512$ \\    $1$\times$1$\times$1, 512$ \end{bmatrix}\times$4 
  & $\begin{bmatrix} $TSM$ \\  $1$\times$1$\times$1, 512$ \\  $1$\times$3$\times$3, 512$ \\    $1$\times$1$\times$1, 512$ \end{bmatrix}\times$4 
  & T$\times$28$\times$28  
  \\
  \hline  
  res$_4$ & $\begin{bmatrix} $1$\times$3$\times$3, 256$ \\    $1$\times$3$\times$3, 256$ \end{bmatrix}\times$2 
  & $\begin{bmatrix} $TSM$\\$1$\times$3$\times$3, 256$ \\    $1$\times$3$\times$3, 256$ \end{bmatrix}\times$2  
  & $\begin{bmatrix} $1$\times$1$\times$1, 1024$ \\ $1$\times$3$\times$3, 1024$ \\    $1$\times$1$\times$1, 1024$ \end{bmatrix}\times$6 
  & $\begin{bmatrix} $TSM$\\  $1$\times$1$\times$1, 1024$ \\ $1$\times$3$\times$3, 1024$ \\    $1$\times$1$\times$1, 1024$ \end{bmatrix}\times$6 
  & T$\times$14$\times$14 
  \\
  \hline  
    res$_5$ & $\begin{bmatrix} $1$\times$3$\times$3, 512$ \\    $1$\times$3$\times$3, 512$ \end{bmatrix}\times$2 
    & $\begin{bmatrix} $TSM$\\$1$\times$3$\times$3, 512$ \\    $1$\times$3$\times$3, 512$ \end{bmatrix}\times$2  
    & $\begin{bmatrix} $1$\times$1$\times$1, 2048$ \\ $1$\times$3$\times$3, 2048$ \\    $1$\times$1$\times$1, 2048$ \end{bmatrix}\times$3 
    & $\begin{bmatrix} $TSM$\\  $1$\times$1$\times$1, 2048$ \\ $1$\times$3$\times$3, 2048$ \\    $1$\times$1$\times$1, 2048$ \end{bmatrix}\times$3 
    & T$\times$7$\times$7
    \\
  \hline  
\multicolumn{5}{c|}{global average pool, FC} & \# of classes\\ 
      \hline
    \end{tabular}
    }
  \end{center}
\end{table}

\begin{table}[t]
\setlength{\tabcolsep}{1pt}
\begin{minipage}{.48\linewidth}
\centering
\begin{center}
\includegraphics[width=0.7\linewidth]{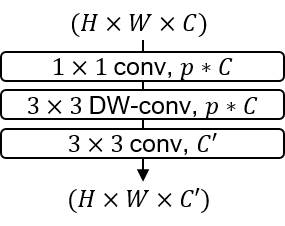}
\end{center}
\captionof{figure}{A bottleneck($p,C'$) module of MobileNet-V2~\cite{sandler2018mobilenetv2}. The module transforms $C$ channels to $C'$ channels with an expansion factor $p$. DW-conv denotes a depth-wise convolution~\cite{howard2017mobilenets}.}   
\label{fig:s1}
\end{minipage}\hfill
\begin{minipage}{.48\linewidth}
\centering
\caption{MobileNet-V2 backbone. Bottleneck modules in Figure~\ref{fig:s1} are applied to the backbone.}
\label{tab:mobile}    
\scalebox{0.9}{
\begin{tabular}{c|c|c}
    \hline
      Layers & MobileNet-V2 & Output size \\
 \hline
      $stage_1$ & 1$\times$7$\times$7, 32, stride 1,2,2 &
      T$\times$112$\times$112  \\
  \hline
      \multirow{2}{*}{$stage_2$}
        & bottleneck(1,16)     
        &\multirow{2}{*}{T$\times$56$\times$56}    
        \\\cline{2-2}& bottleneck(6,24) $\times$ 2
        \\
  \hline
      $stage_3$
        & bottleneck(6,32) $\times$ 3   
        &T$\times$28$\times$28    
        \\
  \hline
      \multirow{2}{*}{$stage_4$}
        & bottleneck(6,64)  $\times$ 4    
        &\multirow{2}{*}{T$\times$14$\times$14}    
        \\\cline{2-2}& bottleneck(6,96) $\times$ 3
        \\
  \hline
      \multirow{3}{*}{$stage_5$}
        & bottleneck(6,160)  $\times$ 3    
        &\multirow{3}{*}{T$\times$7$\times$7}    
        \\\cline{2-2}& bottleneck(6,320)
        \\\cline{2-2}&  1$\times$1$\times$1, 1280, stride 1,1,1    
        \\
    \hline
\multicolumn{2}{c|}{global average pool, FC} & \# of classes\\ 
      \hline
    \end{tabular}
    }
\end{minipage}
\end{table}

\clearpage
\clearpage

\end{document}